\documentclass[runningheads]{llncs}
\usepackage{amsmath, nccmath}
\usepackage{amssymb}
\usepackage{graphicx}
\usepackage{wrapfig}
\usepackage{subfigure}
\usepackage{colortbl}
\usepackage{color}
\usepackage{url}
\usepackage{pdflscape}
\usepackage{rotating}
\usepackage{multirow}
\usepackage{rotating}

\begin{document}
\title{Interpretable Feature Construction\\for Time Series Extrinsic Regression}
%\thanks{Supported by organization x.}}
%
\titlerunning{Feature construction for TSER}
% If the paper title is too long for the running head, you can set
% an abbreviated paper title here
%
 \author{Dominique Gay\inst{1} \and
 Alexis Bondu\inst{2} \and
Vincent Lemaire\inst{2} \and
 Marc Boull\'e\inst{2}}% \and Fabrice Cl\'erot\inst{2}}
 \authorrunning{D.~Gay et al.}
% %First names are abbreviated in the running head.
% %If there are more than two authors, 'et al.' is used.

 \institute{LIM-EA2525, Universit\'e de La R\'eunion, France\\
 \email{dominique.gay@univ-reunion.fr} \and 
 Orange Labs, France\\
 \email{firstname.name@orange.com}}

\maketitle              % typeset the header of the contribution
\begin{abstract}
Supervised learning of time series data has been extensively studied for the case of a categorical target variable. In some application domains, e.g., energy, environment and health monitoring, it occurs that the target variable is numerical and the problem is known as \textit{time series extrinsic regression} (TSER). In the literature, some well-known time series classifiers have been extended for TSER problems. As first benchmarking studies have focused on predictive performance, very little attention has been given to interpretability. To fill this gap, in this paper, we suggest an extension of a Bayesian method for robust and interpretable feature construction and selection in the context of TSER. Our approach exploits a relational way to tackle with TSER: \textit{(i)}, we build various and simple representations of the time series which are stored in a relational data scheme, then, \textit{(ii)}, a propositionalisation technique (based on classical aggregation/selection functions from the relational data field) is applied to build interpretable features from secondary tables to ``flatten'' the data; and \textit{(iii)}, the constructed features are filtered out through a Bayesian Maximum A Posteriori approach. The resulting transformed data can be processed with various existing regressors.
Experimental validation on various benchmark data sets demonstrates the benefits of the suggested approach.

%\keywords{Multivariate Time Series Classification, Feature Selection, Bayesian Modeling, Propositionalisation, Interpretable Models}
\end{abstract}
\section{Introduction}
%Introduction general, exemple concret + graphique\\

Time series analysis has attracted much effort of research in the past decade, driven largely by the wide spread of sensors  and their emerging applications in various domains ranging from medicine to IoT industry. The literature about supervised time series classification is abundant~\cite{BLB+17} and dozens of algorithms have been designed to predict a discrete class label for time series data. However, in some application domains, like sentiment analysis, forecasting, and energy monitoring~\cite{TBP+20a}, the target variable is numeric: e.g., the task of predicting the total energy usage in kWh of a house given historical records of temperature and humidity measurements in rooms and weather measurements. This problem is known as \textit{time series extrinsic regression} (TSER~\cite{TBP+20b}). For an incoming time series $\tau = \left\langle (t_1,X_1), (t_2,X_2), \ldots, (t_m,X_m)\right\rangle$, which is a time-ordered collection of $m$ pairs of time stamps $t_i$ and measurements $X_i\in\mathbb{R}^d$, the goal is to predict the value of a numeric target variable, given a training set of $n$ series, $\mathcal{D} = \{(\tau_1, y_1), (\tau_2, y_2), \ldots, (\tau_n, y_n)\}$, where $y_i\in \mathbb{R}$ are the known target values for series $\tau_i$.

Classical regression algorithms like, e.g., linear regression, regression tree, random forest or support vector regression can deal with TSER, provided that potential multiple dimensions of the input series are concatenated into a single feature vector. Beside $k$ nearest neighbors models using popular distance metrics, like Euclidean distance (ED) and Dynamic Time Warping (DTW), Tan et al.~\cite{TBP+20b} suggest a TSER benchmarking study involving also three recent deep learning approaches~\cite{fawaz2018deep} (FCN, ResNet, InceptionTime~\cite{FLF+20}) and an adaptation of Random Convolutional Kernel Transform (Rocket~\cite{DPW20}) for regression tasks. The first benchmarking study in~\cite{TBP+20b} evaluates 13 TSER algorithms with a focus on predictive performance comparison with root mean squared error (RMSE) as performance measure. As a result, Rocket scores the best mean rank although no significant difference of performance is observed compared with classical regression ensembles like XGBoost~\cite{CG16} or random forest~\cite{Bre01}.

\begin{figure}[htbp!]
\centering
\subfigure[]{\label{scatter}\includegraphics[width=0.49\linewidth]{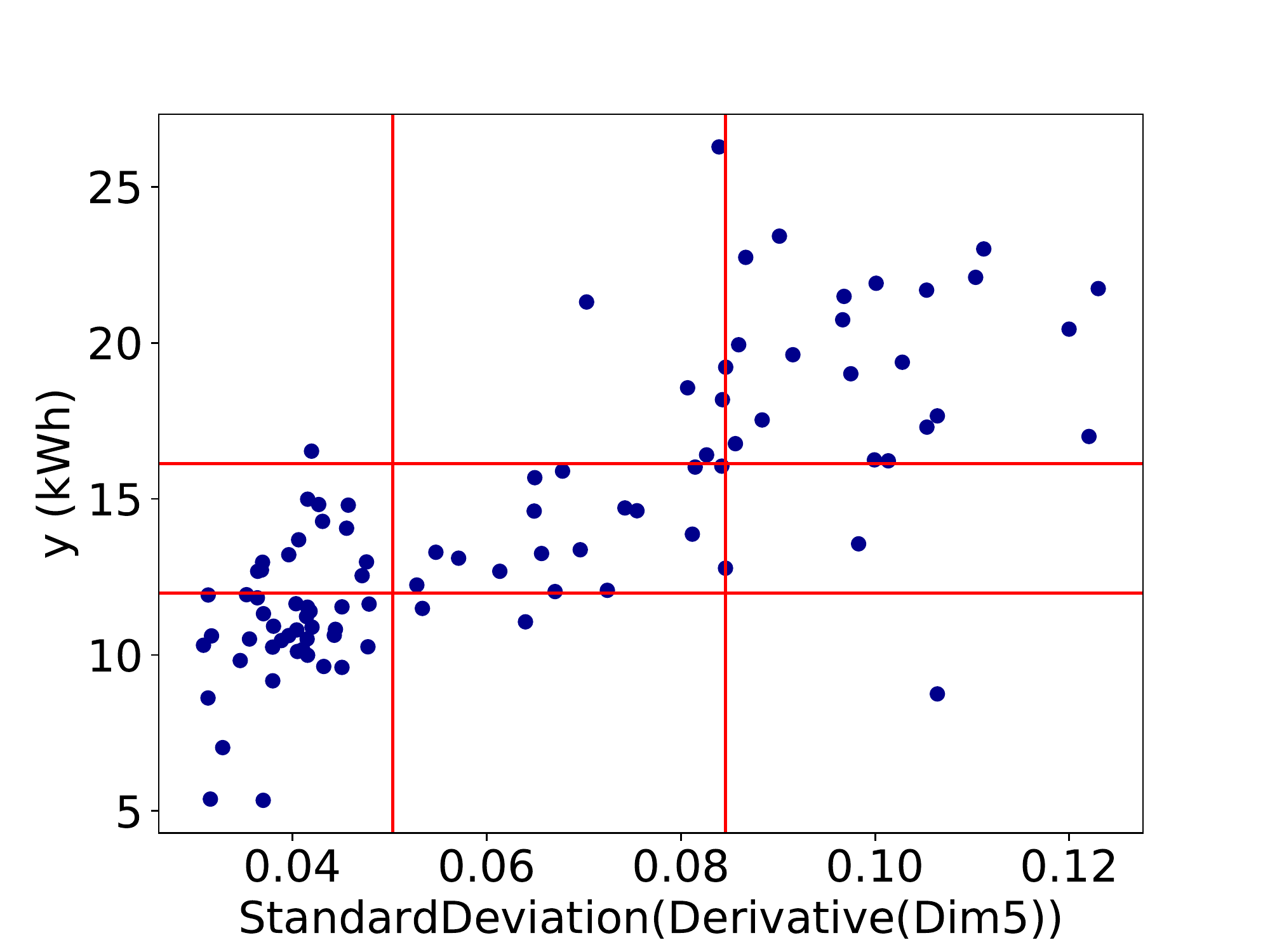}}
%\quad
\subfigure[]{\label{histo}\includegraphics[width=0.49\linewidth]{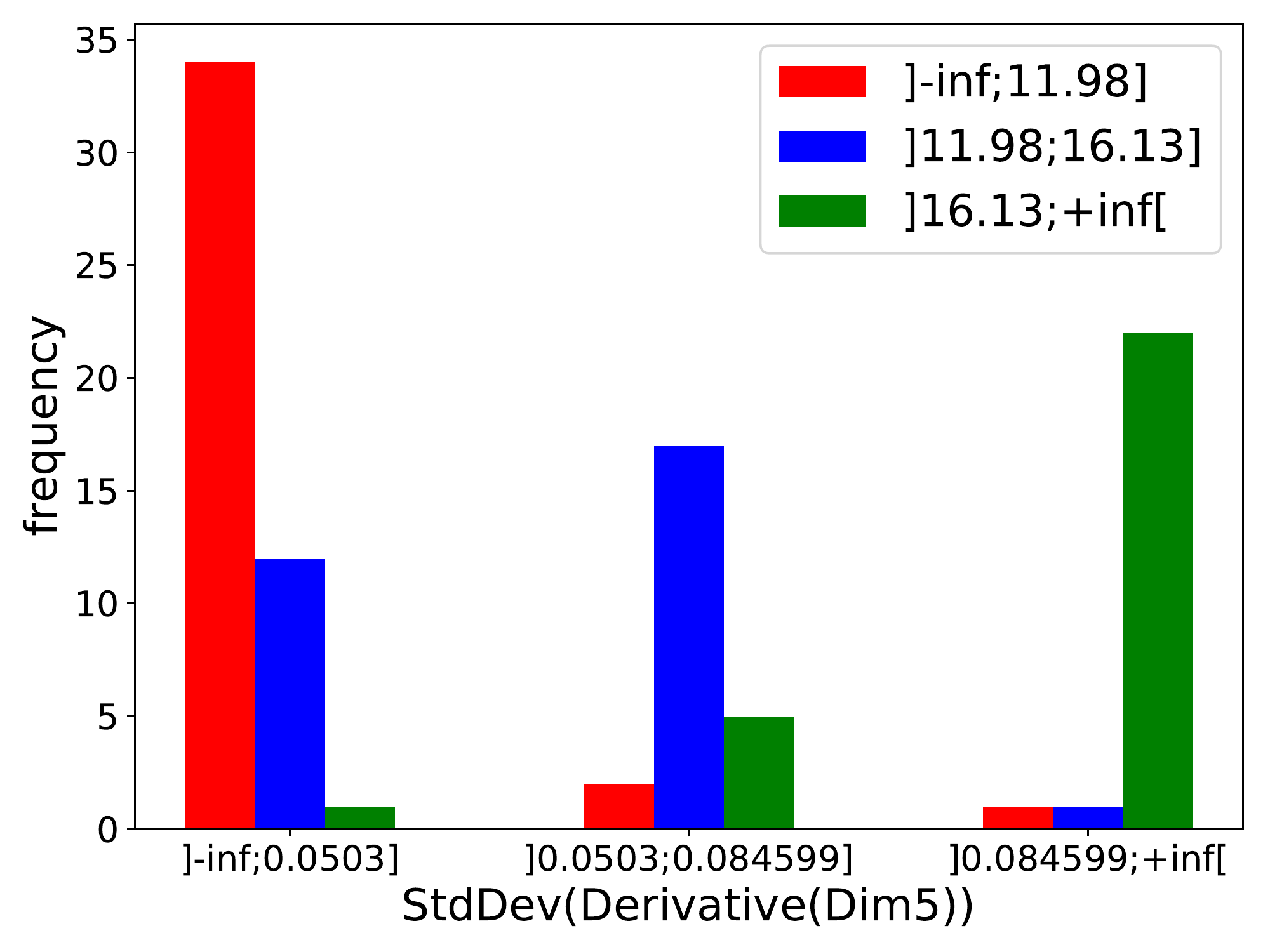}}
\vspace{-0.5cm}
\caption{\footnotesize (a) Scatter plot of $v = StdDev(Derivative(Dim5))$ versus target $y$. One point represents a training time series $\tau_i$. (b) Histograms of frequencies from the contingency table of \textit{intervals of $v$ $\times$ intervals of $y$}} 
\label{fig:motivation}
\vspace{-0.5cm}
\end{figure}

In this paper, we exploit a relational machine learning approach~\cite{BoulleML17} for interpretable feature construction and selection and suggest an extension for TSER problems.
% \begin{wrapfigure}{r}{0.48\textwidth}
% %\begin{minipage}{.48\textwidth}
% %\begin{table}[htbp!]
% 	\centering
% 	\includegraphics[width=.48\textwidth]{img/histo_appliances_energy.png}
% 	\caption{\scriptsize Distibution}% (i.e., the $y$-axis of the gyroscope). }
% 	\label{fig:motiv}
% %\end{table}
% %\end{minipage}
% \end{wrapfigure}
%
As a motivating example, we consider the \textit{AppliancesEnergy} data~\cite{TBP+20a}. The goal is to predict the total energy usage in kWh in a house given 24-dimensional time series recording historical temperature and humidity measurements in 9 rooms in addition to 6 other weather and climate data series. In this context, let us consider \textit{(i)}, the constructed variable {\scriptsize{$v = StdDev(Derivative(Dim5))$}}, i.e., the standard deviation of the derivative transform of dimension 5, and its discretisation into three informative intervals and \textit{(ii)}, the discretisation of the target variable $y$ into three intervals (see Fig.~\ref{fig:motivation}(a)). Plotting frequency histograms in this 2D-grid discretisation (i.e., contingency table of \textit{intervals of $v$ $\times$ intervals of $y$}) directly highlights that variations of measurements related by dimension 5 are characteristic of total power usage and the interpretation is straightforward. Indeed, low $v$ values (below 0.0503) mainly means low power usage (below 11.98), higher $v$ values (above 0.0846) means higher power usage (above 16.13) and in between values of $v$ are characteristic of target interval $]11.98 ; 16.13]$.

%\vspace{-1.1cm}

Our approach, called \texttt{iFx}, brings a methodological contribution to TSER problems as it aims at generalizing the underlying concepts of the above intuitive example to efficiently extract simple and interpretable features as follows: \textit{(i)}, firstly, we transform the original series into multiple representations which are stored in secondary tables as in relational data scheme; \textit{(ii)}, then, informative and robust descriptors are extracted from relational data through propositionalisation and selected using a regularized Bayesian method. Classical regression algorithms can be trained on the obtained flattened data.

The rest of the paper successively presents the main concepts of our approach in Section~\ref{sec:approach}, the experimental validation in Section~\ref{sec:xp} and opens future perspectives after concluding in Section~\ref{sec:conc}.
% %
% As far as we know, MTSC have not yet been approached from a relational data classification point of view. Our approach, called KMTS, brings a methodological contribution to MTSC literature as it generalizes the underlying concepts of the above intuitive example to efficiently extract simple and interpretable features for MTSC, as follows: \textit{(i)}, firstly, we transform the original MTS into multiple representations which are stored
% in secondary tables as in relational data scheme; \textit{(ii)}, then, informative and robust descriptors are extracted from relational data, using a regularized Bayesian propositionalisation method; \textit{(iii)}, thirdly, a selective Na\"{i}ve Bayes classifier is trained on the obtained flattened data.
% %

% The rest of the paper successively presents the main concepts of our KMTS approach in Section~\ref{sec:approach}, the experimental validation in Section~\ref{sec:xp} and opens future perspectives after concluding in Section~\ref{sec:conc}.

%\input{related_work}

%\newpage
\section{TSER via a relational way}
\label{sec:approach}
%
%approach, relational, propositionalisation, feature construction, selection and classification
%

To extract features such as in the illustrative example, our approach is based on \textit{(i)} the computation of multiple yet simple representations of time series, and their storage in a relational data scheme, \textit{(ii)} a recent approach for feature engineering through propositionalisation~\cite{BoulleML17} and its extension for regression problems. In the following, we describe these two steps in order to make the paper self-contained.

\paragraph{\textbf{Multiple representations of series in a relational scheme - }}
Enriching time series data with multiple transformations at the first stages of the learning process has demonstrated a significant enhancement of the predictive performance for the case of supervised classification~\cite{bagnall2012,LTB18,BGL+19,GBL+20}. As transforming time series from the time domain to an alternative data space is a good way for accuracy improvement, we also build six simple transformations commonly used in the literature in addition to the original representation: local derivatives (D) and second-order derivatives (DD), cumulative and double cumulative sums (S and SS), the auto-correlation transform (ACF) and the power spectrum (PS).

To embed all the computed representations in the same data, we use a relational data scheme. The root table is made of two attributes (columns), the series ID and the Class value. Each representation of a dimension of a series is stored in a secondary table in which there are three attributes, the series ID, linked with the the series ID of the root table by a foreign key, the Time attribute (or frequency for PS transform) and the Value attribute (or power for PS transform); thus, each tuple (line) of a secondary table is a single data point. For the introductory Appliances Energy 24-dimensional time series data, the resulting relational scheme is made of $7 \times 24$ secondary tables.

\paragraph{\textbf{Feature construction through propositionalisation - }}Propositionalisation~\cite{DL01,lachiche17} is the natural way to build features from secondary tables. It simply adds columns (variables) containing information extracted from secondary tables to the root table. For TSER data, propositionalisation may generate different aggregate features from various representations of the multiple dimensions.
The introductory variable $v = StdDev(Derivative(Dim5))$, i.e., \textit{``the standard deviation of the derivative transform of dimension 5''} is an example of the type of variables we want to build. It involves an aggregate function (or operator), $StdDev$, the standard deviation and the Value attribute of the table containing the derivative transform of the original fifth dimension that we build in the previous step.

To generalize from this example and build various types of interpretable features while avoiding intractable search space, we suggest propositionalisation through a restricted language, i.e., we will use a finite set of functions. As in a programming language, a function is defined by its name, the list of its operand and its return value and is expressed n the form $fname([operand,\ldots])\rightarrow value$ -- the operands and return value being typed. The operands can be a column of a table or the output of another function, i.e. another feature. Since time series data are inherently numeric the language of functions is made of:
{\footnotesize
\begin{itemize}
    \item well-known and interpretable aggregate functions coming from relation data base domain 
    \begin{itemize}
        \item $Count(Table) \rightarrow Num$ ; count of records in a table
        \item $Mean(Table, NumFeat) \rightarrow Num$ ; mean value of a numerical feature in a table,
        \item $Median(Table, NumFeat) \rightarrow Num$ ; median value,
        \item $Min(Table, NumFeat) \rightarrow Num$ ; min value,
        \item $Max(Table, NumFeat) \rightarrow Num$ ; max value,
        \item $StdDev(Table, NumFeat) \rightarrow Num$ ; standard deviation,
        \item $Sum(Table, NumFeat) \rightarrow Num$ ; sum of values.
    \end{itemize}
    \item $Selection(Table, selectioncriterion) \rightarrow Table$ ; for the selection of records from the table according
to a conjunction of selection terms (membership in a numerical interval, on a column of the operand table or on a feature built from tables related to the operand table). For TSER data, the selection function allows restriction to intervals of timestamp or value in secondary tables.
\end{itemize}}
Let us consider the variable $w = Min(Selection(derivative, 0<timestamp<10), ValueDim3)$, i.e., \textit{the minimum value of the derivative transform of dimension 3 in the time interval $[0;10]$} as an example of the use of the selection function. Here, the $Min$ function is applied on the output of the \textit{selection} function used to select a specific time period.

Given the aforementioned language, to construct a given number $K$ of variables, we use simultaneous random draws to efficiently sample the search space~\cite{BoulleML17}.

\paragraph{\textbf{Feature selection through Bayesian Maximum A Posteriori - }}
The randomized facet of the propositionalisation step does not guarantee that the $K$ aggregate features of the main table are relevant for target variable prediction. All generated features are numerical due to the nature of aggregate functions. In order to select informative ones, we proceed a supervised pre-processing step which is 2D-discretisation, i.e., similarly to the 2D-grid in Fig.~\ref{fig:motivation}, we jointly partition each pair $(v,y)$, where $v$ is  an aggregate feature and $y$ the target variable.

In the Bayesian framework~\cite{HB07}, 2D-discretisation is turned into a model selection problem and solved in Bayesian way through optimization algorithms. According the Maximum A Posteriori (MAP) approach, the best discretisation model $M_{v,y}^*$ is the one that maximizes \textit{the probability of a discretisation model given the input data $D$}, which is:

\begin{equation}
\medmath{
P(M_{v,y}\mid D) \propto P(M_{v,y})\times P(D\mid M_{v,y})
\label{map}}
\end{equation}
	
Switching to negative logarithm refers to information theory and coding lengths. We define a $cost$ criterion, noted $c$: 
\begin{equation}
\medmath{
c(M_{v,y}) = -\log(P(M_{v,y})) -\log(P(D\mid M_{v,y})) = L(M_{v,y}) + L(D | M_{v,y})
\label{bayesCriterion}}
\end{equation}
In terms of information theory, this criterion is interpreted as coding lengths~\cite{Sha01}: the term $L(M_{v,y})$ represents the number of bits used to describe the model and $L(D | M_{v,y})$ represents the number of bits used to encode the target variable with the model, given the model $M_{v,y}$.

The prior $P(M_{v,y})$ and the likelihood $P(D\mid M_{v,y})$ are both computed with the parameters of a specific discretisation which is entirely identified by:
\begin{itemize}
    \item a number of intervals $I$ and $J$ for $v$ and $y$,
    \item a partition of $v$ in intervals, specified on the ranks of the values of $v$,
    \item for each interval $i$ of $v$, the distribution of instances over the intervals $j$ of $y$, specified by $N_{ij}$, the instance counts locally to each interval of $v$.
\end{itemize}
Therefore, according to ~\cite{HB07}, using a prior that exploits the hierarchy of parameters that is uniform at each stage of the hierarchy, allows us to obtain an exact analytical expression of the cost criterion:

\begin{align}
\small
c(M_{v,y}) =\quad & 2\log(N) + \log \binom{N + I - 1}{I - 1} + \sum_{i=1}^{i=I} \log\binom{N_{i.} + J -1}{J - 1}\label{eq:prior}\\
	& + \sum_{i=1}^{i=I} \log\frac{N_{i.}!}{N_{i1}!N_{i2}!\ldots N_{iJ}!} + \sum_{j=1}^{j=J} \log N_{.j}! \label{eq:likelihod}
\end{align}

where $N_{i.}$ is the number of instances in interval $i$ of $v$ and $N_{.j}$ the number of instances in interval $j$ of $y$. The prior part (Eq.~\ref{eq:prior}) of the cost criterion favors simple models with few intervals, and the likelihood part (Eq.~\ref{eq:likelihod}) favors models that fit the data regardless of their complexity.  
Since the magnitude of the cost criterion depends on the size of the data $N$, we define a normalized version, which can be interpreted as a compression rate and is called $level$:
	\begin{equation}
	\medmath{
    level(M_{v,y}) = 1 - \frac{c(M_{v,y})}{c(M_{v,y}^\emptyset)}}
    \end{equation}
where $c(M_{v,y}^\emptyset)$ is the cost of the null model (i.e. when $v$ and $y$ are partitioned into only one interval). The cost of the null model can be deduced from previous formula and is formally $c(M_{v,y}^\emptyset)) = 2\log(N) + \log(N!)$.\\
For example, again for the Appliances Energy data of Fig.~\ref{fig:motivation}, for $N=95$, $c(M_{v,y}^\emptyset))\simeq505$ ; $c(M_{v,y}) \simeq 13 + 12 + 27 + 72 + 349 = 473$ ; thus the level of $v$ is positive as  $level(M_{v,y})\simeq 0.0614$.

A variable $v$ whose discretisation model $M_{v,y}$ obtains a positive level value will be considered as informative whereas negative level value indicates spurious variables. Indeed, with negative level, a discretisation model $M_{v,y}$ is less probable than the null model, thus irrelevant for the regression task. When $0<level(M_{v,y})<1$, we reach the most probable models that highlight a correlation between $v$ and $y$.
To reach those models, we use classical greedy bottom-up algorithms~\cite{BoulleML06} that allows to efficiently find the most probable model given the input data in $O(N\log N)$ time complexity, where $N$ is the number of time series.

At the end of our iFx process, we obtain a classical tabular data format, i.e., the series are now described by informative and interpretable features selected among the $K$ variables extracted from multiple representations of original series. And on-the-shelf regressions algorithms can be trained.

As an example, for Appliances Energy data, one can find below the 17 selected variables (with positive $level$ values) among the $K=1000$ extracted, and their corresponding optimal number of intervals $J$ and $I$ for target variable $y$ and $v$.

\begin{center}
{\scriptsize
\begin{tabular}{lrcc}
\hline
Feature $v$ & $level(v)$ & \#TargetIntervals & \#$v$Intervals\\
\hline
$StdDev(TS5D.Value5D)$	& 0.0485	& 3	& 3\\
$StdDev(TS6DD.Value6DD)$	& 0.0395	& 3	& 2\\
$Max(TS6DD.Value6DD)$	& 0.0360	& 2	& 2\\
$StdDev(TS6D.Value6D)$	& 0.0345	& 3	& 2\\
$Max(TS5D.Value5D)$	& 0.0302	& 2	& 2\\
$StdDev(TS5DD.Value5DD)$	& 0.0299	& 2	& 2\\
$Min(TS6DD.Value6DD)$	& 0.0275	& 3	& 2\\
$StdDev(TS5.Value5)$	& 0.0253	& 2	& 2\\
$Mean(TS5PS.Value5PS)$	& 0.0242	& 2	& 2\\
$Sum(TS5PS.Value5PS)$	& 0.0242	& 2	& 2\\
$Max(TS6D.Value6D)$	& 0.0213	& 2	& 2\\
$Min(TS6D.Value6D)$ & 0.0211	& 2	& 2\\
$Max(TS5DD.Value5DD)$	& 0.0207	& 2	& 2\\
$Max(TS5PS.Value5PS)$	& 0.0171	& 2	& 2\\
$StdDev(TS5PS.Value5PS)$	& 0.0171	& 2	& 2\\
$Min(TS5D.Value5D)$	& 0.0162	& 2	& 2\\
$Min(TS5DD.Value5DD)$	& 0.0077	& 3	& 2\\ 
\hline
\end{tabular}
}
\end{center}

\section{Experimental validation}
\label{sec:xp}
The experimental evaluation of our approach iFx are performed to discuss the following questions:
\begin{itemize}
    \item[$Q_1$] Concerning iFx, how does the predictive performance evolve w.r.t. the number generated features ? How many relevant features are selected? Are there preferred dimensions/representations for feature selection? And what about the time efficiency of the whole process?
    \item[$Q_2$] Are the performance of iFx comparable with current TSER methods?
\end{itemize}
\paragraph{\textbf{Experimental protocol \& data sets - }} Tan et al.~\cite{TBP+20a} has recently released 19 TSER data sets that are publicly available. The repository exhibits a large variety of TSER application domains with various numbers of dimensions and series' lengths. Predefined train/test sets are provided and we used it per se. As end classifiers, we use Python implementations of regression tree (RT) and random forest (RF) from scikit-learn library~\cite{PVG+11}, Gradient Boosting Trees\footnote{\url{https://xgboost.readthedocs.io/en/latest/python/index.html}} (XGB) and the C++ implementation of the Selective Naive Bayes (SNB)~\cite{BoulleJMLR07} since iFx and SNB is part of the same tool\footnote{\url{http://www.khiops.com} \emph{(available as a shareware for research purpose)}}. All implemntations are used with default parameters except for RF and XGB for which the number of trees is set to 100. Notice that SNB is parameter-free. The Root Mean Squared Error (RMSE) serves as the predictive performance measure. All experiments are run under a laptop with Ubuntu 20.04 using an Intel Core i7-5500U CPU@ 2.40GHz x4 and 12Go RAM.
%% OK %%\textcolor{magenta}{(VL: un peu court pour les paramètres : RF paramètres par défaut ? XGboost paramètres par défaut ? si oui dans les deux cas les donner)}
% \begin{itemize}
%     \item XP k 1..1000 for DTreg rmse
%     \item running times
%     \item transform distribution (selection), ratio generated informative
%     \item robustness
%     \item iFx1000 + XGBoost/RFR/DTreg
%     \item best iFx vs sota CDC
% \end{itemize}

%\vspace{-0.5cm}
\begin{table}[htbp!]
\scriptsize
\centering
%\rotatebox{90}{
\begin{tabular}{|l|rrrr|rr|rr|r|}
\hline
Data                       & RT      & 10-RT & 100-RT & 1000-RT & RF     & 1000-RF & XGB     & 1000-XGB & 1000-SNB\\
\hline
%Data        & RT      & IFX10-RT & IFX500-RT & IFX1000-RT & RF     & IFX1000-RF & XGB     & IFX1000-XGB & IFX1000-SNB \\
AE          & 5.903   & 3.455    & 4.158     & 2.7        & 3.397  & 1.999      & 4.024   & 2.202       & 2.463       \\
HPC1        & 429.067 & 473.538  & 143.29    & 118.692    & 256.14 & 58.729     & 278.086 & 107.817     & 56.431      \\
HPC2        & 58.754  & 59.043   & 58.99     & 46.874     & 46.941 & 36.51      & 48.571  & 36.813      & 38.529      \\
BC          & 0.808   & 9.978    & 5.545     & 4.39       & 0.838  & 3.187      & 0.607   & 4.429       & 4.119       \\
BPM10       & 140.825 & 142.955  & 136.61    & 127.906    & 94.759 & 94.596     & 95.542  & 96.486      & 109.163     \\
BPM25       & 85.091  & 95.461   & 85.551    & 87.066     & 62.787 & 63.812     & 62.325  & 62.891      & 75.799      \\
LFMC        & 62.169  & 58.499   & 59.418    & 61.764     & 44.589 & 45.477     & 47.549  & 48.109      & 44.551      \\
FM1         & 0.022   & 0.017    & 0.009     & 0.01       & 0.016  & 0.007      & 0.017   & 0.008       & 0.009       \\
FM2         & 0.019   & 0.008    & 0.007     & 0.007      & 0.015  & 0.006      & 0.018   & 0.008       & 0.015       \\
FM3         & 0.021   & 0.017    & 0.01      & 0.009      & 0.021  & 0.008      & 0.02    & 0.007       & 0.009       \\
AR          & 14.62   & 11.46    & 51.578    & 43.011     & 8.526  & 14.873     & 8.918   & 12.64       & 8.266       \\
PPGDalia    & 24.964  & 24.165   & 24.899    & 22.008     & 17.487 & 16.083     & 16.144  & 15.985      & 16.819      \\
IEEEPPG     & 39.75   & 49.169   & 41.041    & 43.031     & 32.11  & 38.51      & 31.716  & 37.931      & 41.875      \\
BIDMC32HR   & 19.233  & 19.994   & 18.013    & 15.116     & 15.069 & 17.085     & 13.524  & 15.112      & 13.729      \\
BIDMC32RR   & 4.709   & 5.888    & 4.728     & 5.377      & 4.362  & 4.511      & 4.313   & 4.473       & 3.910       \\
BIDMC32SpO2 & 4.924   & 5.03     & 5.273     & 4.825      & 4.555  & 4.382      & 4.538   & 4.604       & 4.961       \\
NHS         & 0.192   & 0.142    & 0.142     & 0.142      & 0.148  & 0.142      & 0.144   & 0.142       & 0.142       \\
NTS         & 0.186   & 0.14     & 0.177     & 0.183      & 0.143  & 0.143      & 0.14    & 0.14        & 0.138       \\
C3M         & 0.05    & 0.053    & 0.061     & 0.058      & 0.043  & 0.045      & 0.051   & 0.053       & 0.045    \\
\hline
Win vs Orig & - & 9 & 11 & 13 & - & 11 & - & 8 & -\\
\hline
\end{tabular}
%}
\caption{RMSE results of our iFx method with $K=10, 100, 1000$ and with a regression tree (RT), random forest (RF), gradient boosting trees (XGB) and Selective Naive Bayes (SNB) as end regressors.}
\label{tab:results}
\vspace{-8mm}
\end{table}

%\vspace{-1cm}
\paragraph{\textbf{Performance evolution w.r.t. the number of features - }} We study the predictive performance evolution w.r.t.  $K$, the number of extracted features on the 19 data sets~\cite{TBP+20a}. For this experiment, we use a simple DecisionTreeRegressor from scikit-learn library~\cite{PVG+11} as the end regressor. In Table~\ref{tab:results}, we report RMSE results of our approach for increasing $K=10, 100, 1000$. As expected, adding more informative features generally brings better predictive performance. Similar results are observed for RF but due to text width, we only reports results for $K=1000$.

With $K=1000$, the regression tree iFx1000-RT achieves better RMSE results than regression tree on original data for 13/19 data sets. Thus, using the new representation induced by iFx improves the predictive performance of the regression tree. The same observations stand for RF (11/19), and surprisingly, it is not the case for XGB (only 8/19). Notice also that iFx1000-RF seems to be the best combination as, in terms of Win-Tie-Loss, it scores 17-1-1 vs iFx1000-RT, 10-1-8 vs iFx1000-XGB and 11-2-6 vs iFx1000-SNB.
%\textcolor{magenta}{(VL: overfitting???)} je ne sais pas je t'avoue je ne comprends pas pourquoi avec XGB ce n'est pas top.
%%OK c'est vrai mais a la fin c'est le SNB le mieux classé... \textcolor{magenta}{(VL: not known as good regressor since it not solve the RMSE problem)}. 

\begin{figure}[htbp!]
    \centering
    %\vspace{-0.5cm}
    \includegraphics[width=.6\textwidth]{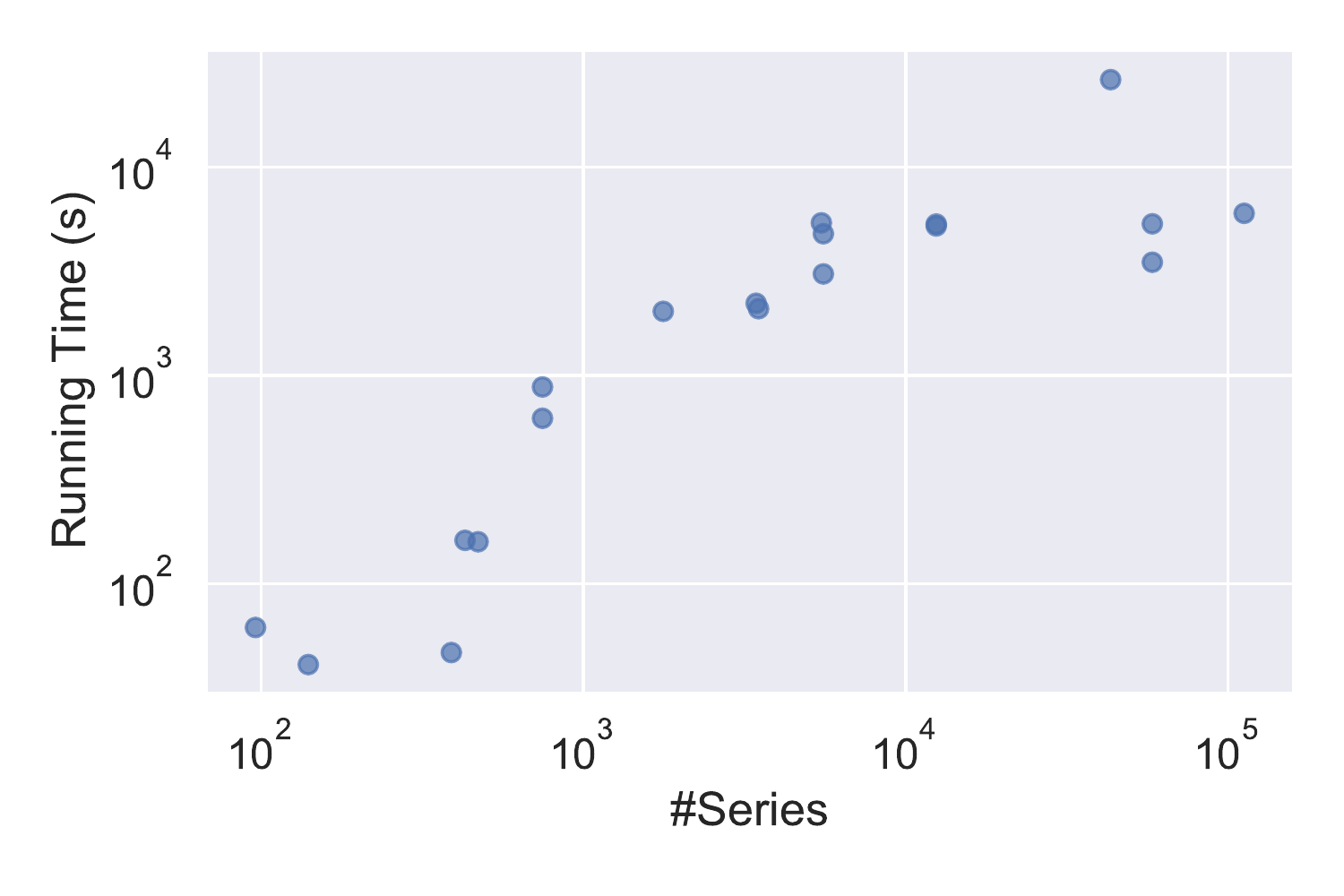}
    \caption{iFx1000 -  Running time in seconds w.r.t. the number of training series for $K=1000$.}
    \label{fig:runningtime}
    %\vspace{-8mm}
\end{figure}

While more features give better predictive performance, it also means higher computational time. As the time complexity of the greedy bottom-up optimisation algorithms, that filter informative features, is supra-linear, then for a fixed $K$, the overall time time complexity is $\mathcal{O}(K.N log(N))$, where $N$ is the number of training series. In practice, with a small computational time overhead to compute the 7 representations, iFx is time-efficient as shown in Figure~\ref{fig:runningtime}. For data sets with less than 1000 series, 1000 seconds are enough and for data with up to 100000 series it runs in 10000 seconds; except for the PPGDalia data which demands 46000 seconds ($\simeq7h)$. We conjecture that this particular high computational time is due to the high precision of the values of the series' data and target variable, that leads to a huge search space for 2D-discretisation model optimisation. Finally, for $K=1000$, iFx process the whole benchmark in 20 hours.

%\begin{wrapfigure}{r}{0.48\textwidth} 
%\end{wrapfigure}

\begin{figure}
    \centering
    \includegraphics[width=\textwidth]{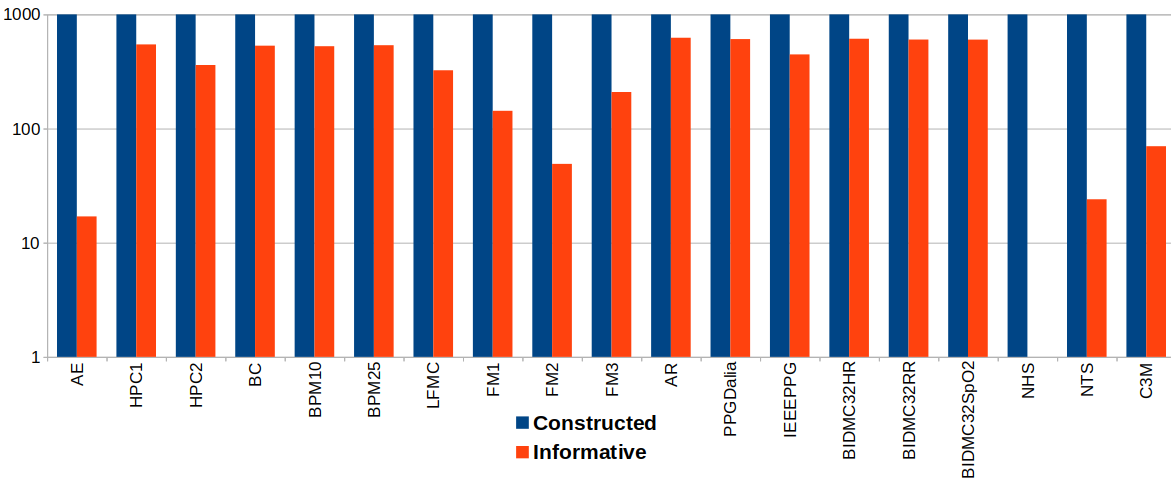}
    %\vspace{-0.4cm}
    \caption{Number of informative features (with positive level) among the $K=1000$ generated features, for each data set.}
    \label{fig:att_dist_infoselec}
\end{figure}

\paragraph{\textbf{Distribution of selected features and representations - }}
As the iFx's feature extraction step by propositionalisation is randomized, we study the proportion of selected informative features (with positive $level$) among the $K$ extracted for each data set in Figure~\ref{fig:att_dist_infoselec}. For most of the data sets, hundreds of the aggregate features are considered as informative, except for Appliances Energy, Flood Modeling 2, News Title Sentiment and Covid3Month. Notice the particular case of News Headline Sentiment (NHS) data, for which none of the 1000 features is informative (i.e., negative $level$ value). We conjecture that neither the present language of aggregate functions nor the transforms are adequate for predictive modeling of NHS data. In this case of no relevant feature, the default way to predict the target value is to predict the mean of the training target values. For NHS data, the default prediction leads to a RMSE value of 0.142 (see Tab.~\ref{tab:results}) on test set and we remark that this is the best score over all contenders in~\cite{TBP+20b}.\\

\begin{figure}
    \centering
    \includegraphics[width=\textwidth]{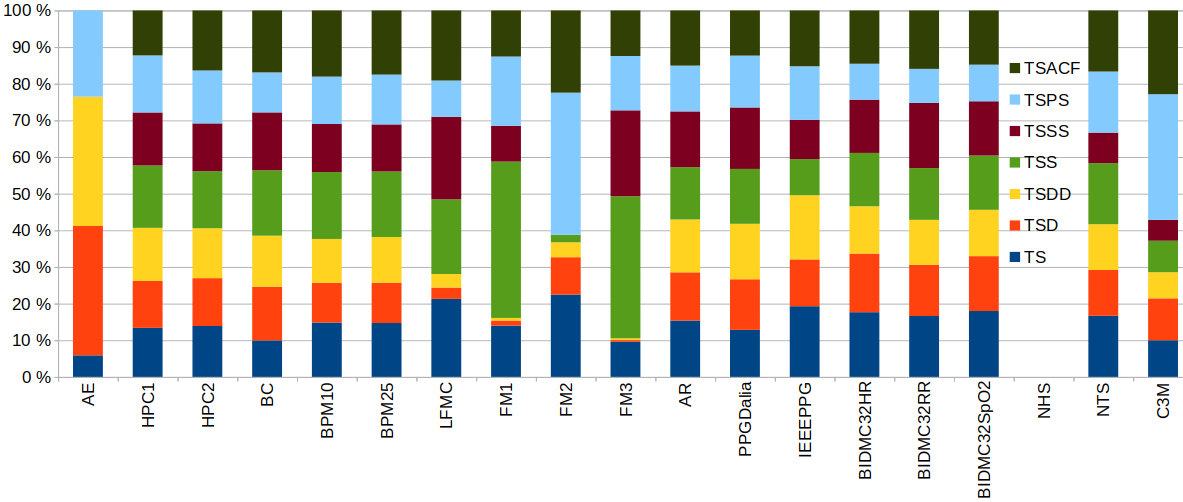}
    %\vspace{-0.4cm}
    \caption{Relative distribution of the seven representations among features selected through iFx (with positive $level$) for each data set.}
    %\vspace{-0.6cm}
    \label{fig:att_dist_rep}
    \vspace{-8mm}
\end{figure}

In Figure~\ref{fig:att_dist_rep}, we study the relative distribution of the selected features into the seven representations for each data set. In most cases, all the seven representations are present in the selected features. A few exceptions stand: for Appliances Energy, S, SS and ACF transformations are of no use for iFx; for Flood Modeling 2, SS transform does not produce any informative attribute and for Live Fuel Moisture Content, Flood Modeling 1 \& 3, very few interesting attributes are coming from D and DD transforms. Now looking at the distribution of the selected features over the dimensions of the series (in the multivariate cases), for most of the data sets, all the dimensions are involved in the selected variables, except for Appliances Energy and News Title Sentiment data where for 22/24 (resp. 2/3) dimensions, no informative attribute has been found.

As for some cases, depending on the data at hand, the transforms of dimensions end up with no informative attributes, there is a potential for further investigations to identify the dimensions and their transforms that are relevant for our language of functions, and then focus the search on them. We postpone this idea for future work.

%\vspace{-1.3cm}

\begin{figure}[htbp!]
    \centering
    \includegraphics[width=.48\textwidth]{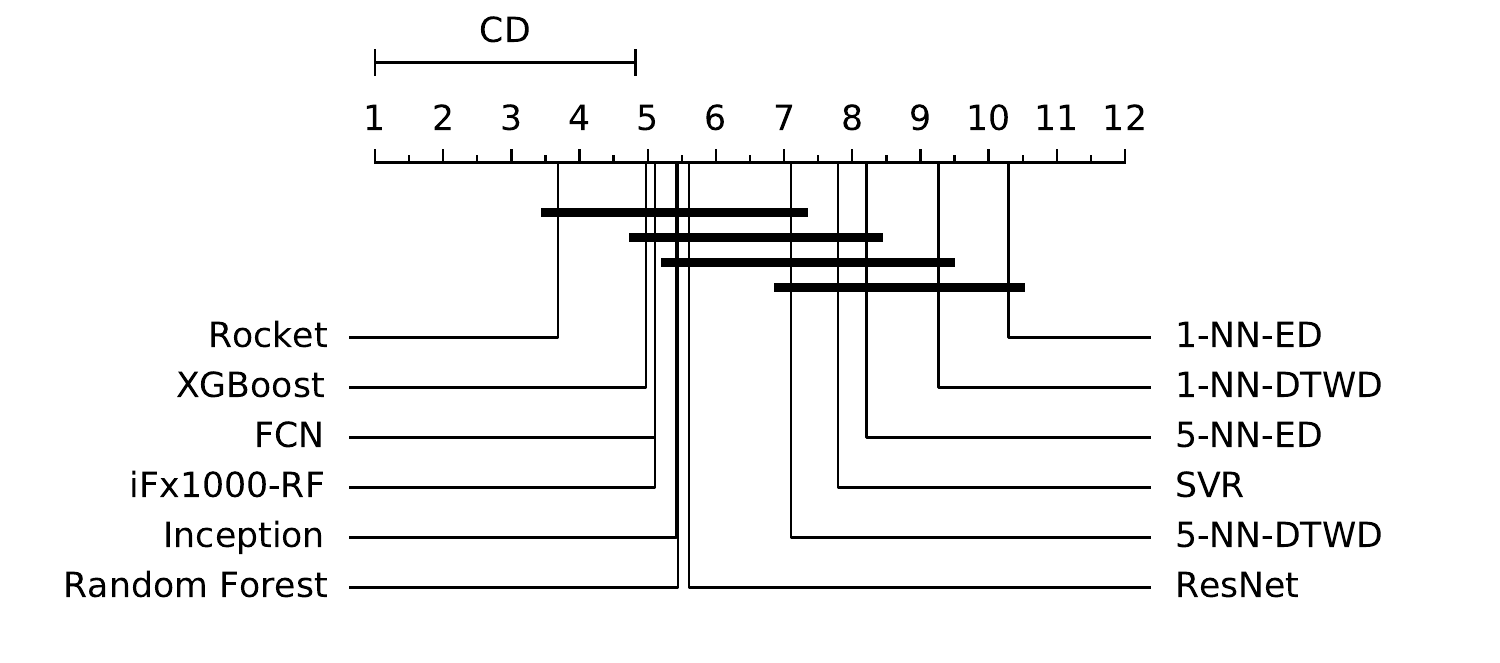}
    \includegraphics[width=.48\textwidth]{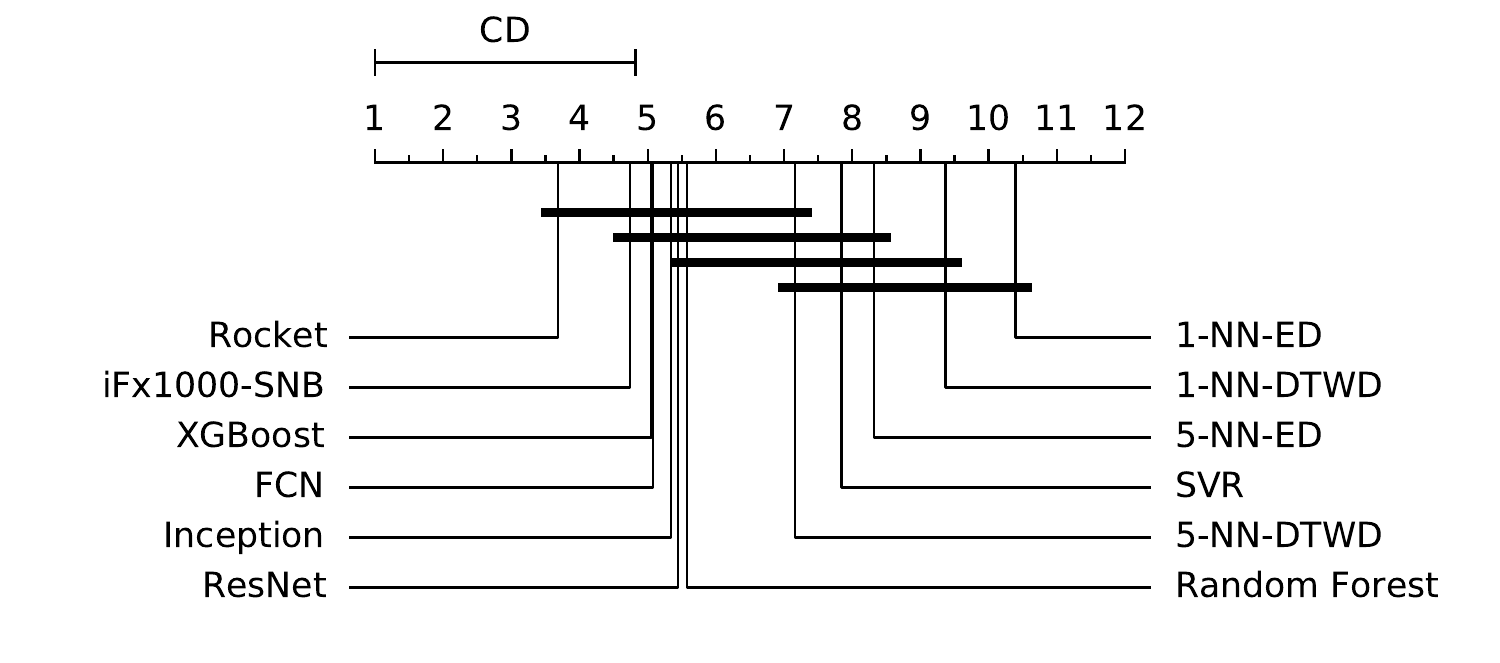}\\
    \includegraphics[width=.48\textwidth]{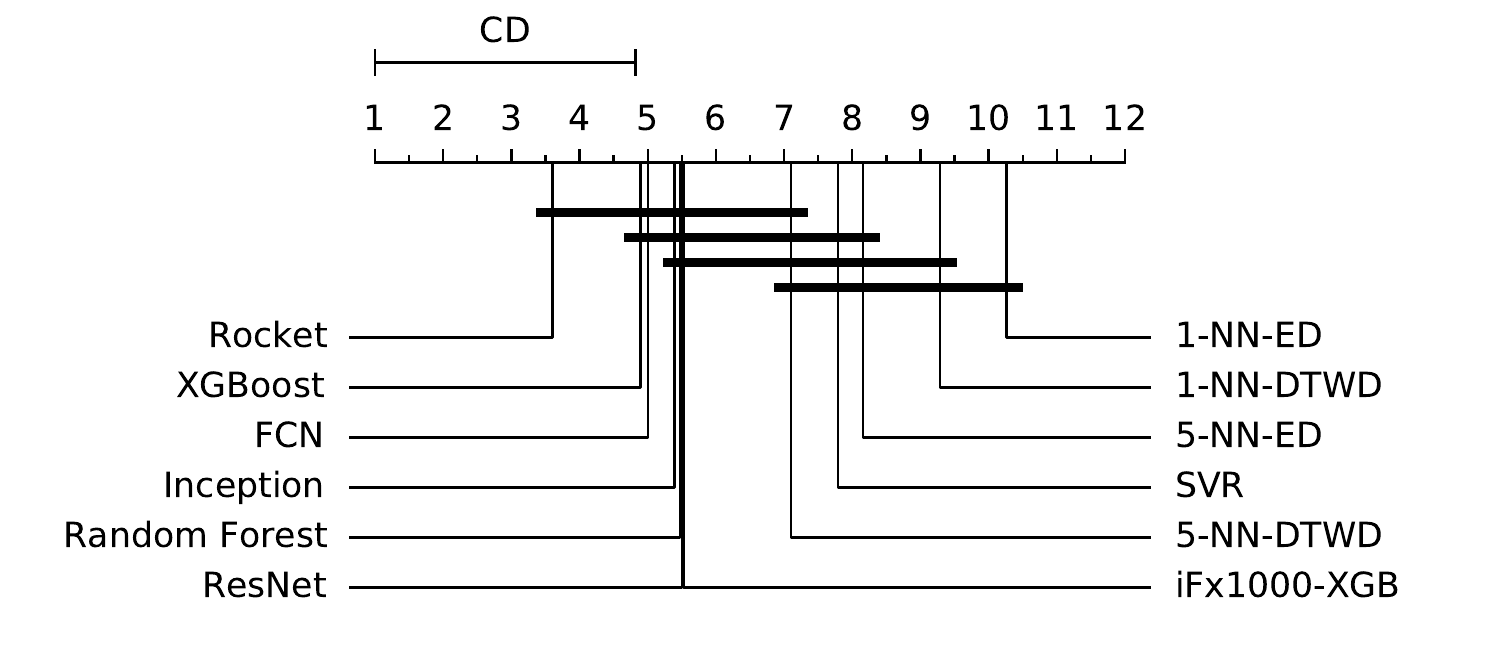}
    \includegraphics[width=.48\textwidth]{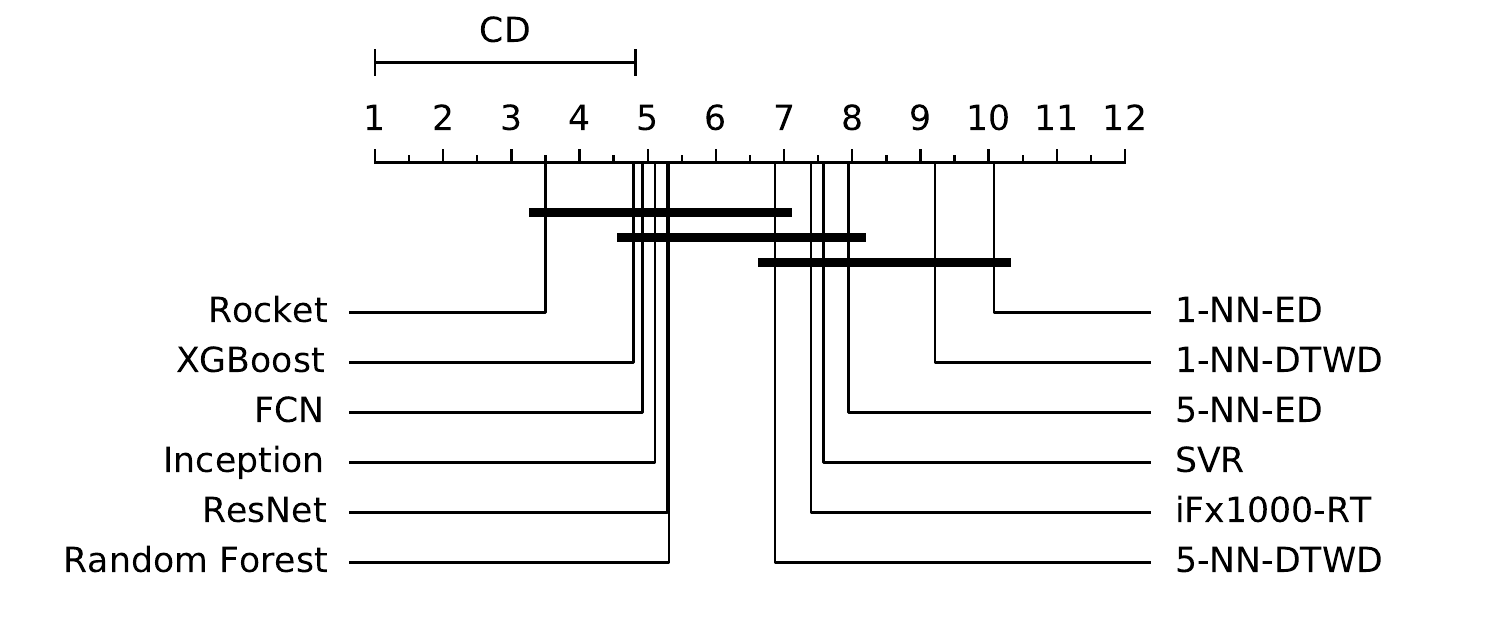}
    \caption{Critical difference diagrams for iFx1000-RF, iFx1000-SNB, iFx1000-XGB and iFx1000-RT versus state-of-the-art regression methods~\cite{TBP+20b}.}
    \label{fig:cd}
    \vspace{-8mm}
\end{figure}

\paragraph{\textbf{Predictive performance comparison with state-of-the-art - }}
In order to compare the predictive performance of iFx with state-of-the-art methods, we use the RMSE results of the contenders from Tan et al.'s benchmarking study~\cite{TBP+20b} and integrate ours for comparison. In Fig.~\ref{fig:cd}, with the critical difference diagrams~\cite{Demsar2006} stemming from Friedman test with post-hoc Nemenyi test, we compare the predictive performance of ifX1000 ended with previous regressors with other contenders. 

Whereas no method is significantly singled out, we observe that Rocket still scores the best mean rank. iFx1000-SNB reaches the second place (Fig.~\ref{fig:cd} (up right)) while iFx1000-RF takes the 4th place (Fig.~\ref{fig:cd} (up left)). These two approaches are then comparable to the best state-of-the-art TSER methods while XGBoost seems to not benefit from our feature engineering method (Fig.~\ref{fig:cd} (down left)).

On the other hand (Fig.~\ref{fig:cd} (down right)), using a single regression tree, the combination iFx1000-RT is 8th with a mean rank of about 7.39 and a significant loss of performance is observed compared with the best ranked approach Rocket. Thus, iFx1000-RT is the most interpretable model of our study but, here, the interpretability is at the cost of performance loss.

%\begin{wrapfigure}[10]{r}{0.5\textwidth}
%\textcolor{magenta}{(VL: Why 3 CD??? , il y aurait aussi la place de mettre la RMSE sur chaque datset des 3, 4 meilleurs?)}
%\end{wrapfigure}
%

\paragraph{\textbf{About the robustness of iFx - }} The $level$ criterion at the core of our feature selection approach is a regularized Bayesian criterion that bets on trade-off between the complexity of a feature and the precision of its contained information. As an additional empirical evidence of the robustness of the $level$ and the good foundation of our approach, we proceed the following experiments. For the Appliances Energy data, we randomly permute the target values in the training set, then we run iFx for $K=10, 100, 1000, 10000, 100000$. As a result, there is strictly no variable with positive $level$ value, i.e., no informative variable.

\section{Conclusion \& Perspectives}
\label{sec:conc}
%Opening on FEARS and beyond
Our methodological contribution, iFx, explores a relational way for TSER problems. iFx is efficient and effective as it is capable of extracting and selecting and interpretable features from simple representations of original time series. Learning classical regression techniques on the new feature set generally results in better predictive performance than with the raw original time series. While Rocket still scores the best mean rank on the TSER benchmark, iFx is comparable with state-of-the-art TSER methods, depending on the end regressor at use.

As future work, we envision two ways of improvement for iFx: \textit{(i)}, enriching the language of functions used in the propositionalisation step; \textit{(ii)}, exploring the space of series dimensions and representations with a feed-forward/feed-backward strategy to reduce the field of extraction of aggregate features to the relevant representations and dimensions.

\bibliographystyle{splncs04}
\bibliography{biblio}

%\newpage
%\input{tabres}

%
%\begin{thebibliography}{8}
%\bibitem{ref_article1}
%Author, F.: Article title. Journal \textbf{2}(5), 99--110 (2016)
%
%\bibitem{ref_lncs1}
%Author, F., Author, S.: Title of a proceedings paper. In: Editor,
%F., Editor, S. (eds.) CONFERENCE 2016, LNCS, vol. 9999, pp. 1--13.
%Springer, Heidelberg (2016). \doi{10.10007/1234567890}
%
%\bibitem{ref_book1}
%Author, F., Author, S., Author, T.: Book title. 2nd edn. Publisher,
%Location (1999)
%
%\bibitem{ref_proc1}
%Author, A.-B.: Contribution title. In: 9th International Proceedings
%on Proceedings, pp. 1--2. Publisher, Location (2010)
%
%\bibitem{ref_url1}
%LNCS Homepage, \url{http://www.springer.com/lncs}. Last accessed 4
%Oct 2017
%\end{thebibliography}
\end{document}